\documentclass{article}
\usepackage{spconf,amsmath,graphicx}
\usepackage[tight,footnotesize]{subfigure}
\usepackage{caption}
\usepackage{epstopdf}
\usepackage{hyperref}
\hypersetup{
	colorlinks   = true, 
	urlcolor     = green, 
	linkcolor    = blue, 
	citecolor   = blue 
}
\usepackage{url}
\AppendGraphicsExtensions{.tif}

\title{WEPSAM: Weakly Pre-Learnt Saliency Model}
\name{Avisek Lahiri$^1$ ~~ Sourya Roy$^2$\sthanks{Shares equal contribution with 1$^{st}$ author.} ~~ Anirban Santara$^3$ ~~ Pabitra Mitra$^3$ ~~ Prabir Kumar Biswas$^1$}
\vspace{-5mm}\address{\small $^1$ Dept. of Electronics and Electrical Communication Engineering, IIT Kharagpur, India \\ \small $^2$Dept. of Instrumentation and Electronics Engineering, Jadavpur University, India\\ \small $^3$Dept. of Computer Science Engineering Engineering, IIT Kharagpur, India}
\begin{document}
%
\maketitle
\begin{abstract}
Visual saliency detection tries to mimic human vision psychology which concentrates on sparse, important areas in natural image.  Saliency prediction research has been traditionally based on low level features such as contrast, edge, etc.  Recent thrust in saliency prediction research is to learn high level semantics using ground truth eye fixation datasets. In this paper we present, WEPSAM : Weakly Pre-Learnt Saliency Model as a pioneering effort of using domain specific pre-learing on ImageNet  for saliency prediction using a light weight CNN architecture. The paper proposes a two step hierarchical learning, in which the first step is  to develop a framework for weakly pre-training on a large scale dataset such as ImageNet which is void of human eye fixation maps. The second step refines the pre-trained model on a limited set of ground truth fixations. Analysis of  loss on iSUN and SALICON datasets reveal that pre-trained network converges much faster compared to randomly initialized network. WEPSAM also outperforms some recent state-of-the-art saliency prediction models  on the challenging MIT300 dataset.
\end{abstract}
\section{Introduction}
\label{sec:intro}

When any scene is presented to the human visual system, it rapidly summarizes it through eye fixation on salient regions of the scene. Visual attention, or more particularly selective visual attention is the main reason behind this phenomenon.  For more than a decade, researchers are trying to develop computational models of selective attention as it’s modeling has numerous important applications across different fields like computer vision, robotics etc. \cite{reid}\cite{robot2}. Many methods of saliency detection have been reported in existing literature and they can be broadly categorized into two groups: low level or bottom-up methods and learning based methods. Low level methods generally seek inspirations from biological processes. Most of the models from this category follow a general pipeline which was first proposed by the seminal work of Itti et al. \cite{Ittikoch}. The authors extracted low level features such as color, orientation, texture etc., from images, computed feature specific saliency maps and finally integrated these to produce master saliency map. ‘Center-surround difference operator is usually employed to construct feature-specific saliency maps. Gao et al. \cite{gao} also compared center and surround features, using KL-Divergence in order to measure distinctness of a specific pixel and subsequently its saliency. Bruce and Tsotsos \cite{Bruce} conjectured salient regions contain maximum ‘self-information’ relative to their surroundings. Seo and Milanfar \cite{Seo} proposed a local ‘self-resemblance’ mechanism based saliency model. Among more recent bottom-up approaches, Murray et al. \cite{Murray} modeled saliency from a color space perspective. Holzbach and Cheng \cite{Holzbach} proposed a method which predicts saliency via calculating dissimilarity between multiple sampling templates. Goferman et al. \cite{context} also exploited mainly low level features for saliency detection; however their model also incorporates face detector for high level feature detection.

Recently, machine learning based approaches have gained popularity because in addition to the low level features, these models also take high level contextual and semantic features into account. As high level features play an important role in driving visual attention, learning based models generally performs better. Judd et al. \cite{Judd} trained a SVM (support vector machine) classifier based model directly from human eye tracking data by utilizing hand crafted low, mid and high level features. Vig et al. \cite{Vig} also projected a similar SVM based algorithm but instead of using hand-tuned features their model learns the ‘optimal saliency features’ automatically from the human eye fixation data. Kavak et al. \cite{kavak} proposed a multiple kernel based learning approach to saliency detection. 
\begin{figure*}[!t]
	\centering
	\includegraphics[scale=0.5]{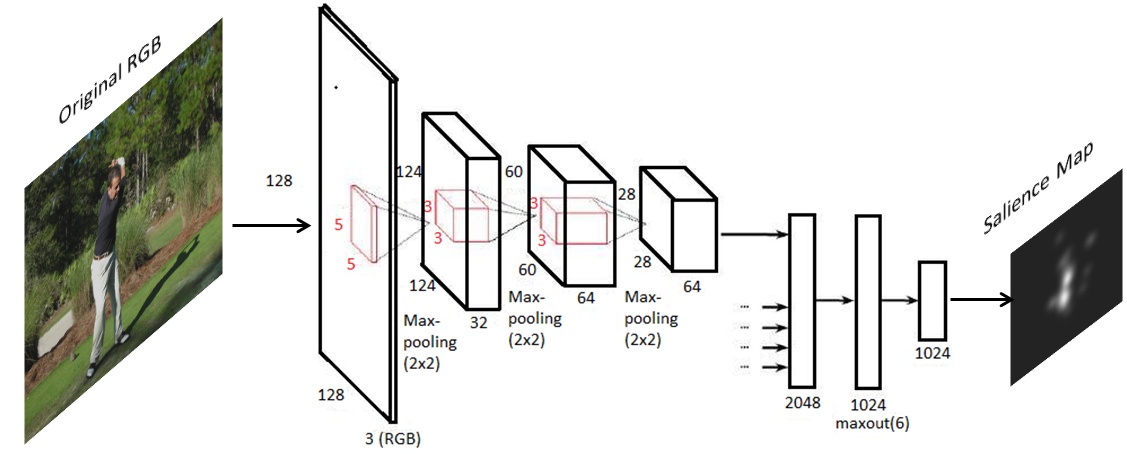}
	\caption{The CNN architecture used for visual saliency prediction.}
	\label{fig_juntingnet_diag}
\end{figure*}

In this paper, we propose an end to end convolutional neural network based model, WEPSAM, for accurate saliency detection. It is a well-known fact that convolutional neural networks (CNN) are very powerful learning systems. From semantic segmentation to object recognition, CNN based models have achieved state of the art performances in a wide range of computer vision tasks. However one major drawback associated with convolutional nets is that their performance critically depends on the size of the dataset. Often large scale datasets, required for proper training of convolutional nets, are not available. To tackle this problem, we introduce a ‘weak data’ driven pre-training paradigm which proves to be a simple but effective solution. The main objective of our work is not to endorse any particular CNN architecture, but rather to present a new training scheme which can help us to train a CNN much faster (compared to a randomly initialized network) for tasks such asIt  saliency prediction where ground truth data is scarce.

The primary contributions of our paper are as follows:
\begin{itemize}
\item To the best of our knowledge, this is the first ever attempt of utilizing ImageNet data for weakly pre-training a CNN. Previous models \cite{salicon,ayush} have attempted to pre-train on ImageNet for object recognition task, but it is more prudent to pre-train a model for domain specific task of saliency prediction. The paper thus opens up a neoteric horizon of effectively leveraging enormous image datasets for visual saliency prediction.
\item  Pre-trained model is then fine-tuned on the actual ground truth fixations. We show that rate of decay of squared error loss of WEPSAM is much faster compared to a randomly initialized CNN network.
\item  We compare our model on the challenging MIT300 dataset with recent state-of-the-art methods on five popularly used metric for saliency prediction task. 
\end{itemize}

\section{CNN Architecture and PARAMETERS}
In this section we briefly describe the CNN architecture used for both pre-training and fine tuning stage. We wish to reiterate that the purpose of this work is not propose or use a very deep CNN architecture, but to study the feasibility of leveraging domain specific pre-training for saliency prediction. We use a shallow CNN with only 5 layers inspired from \cite{juntingnet} with subtle modifications. The network is shown in Fig. \ref{fig_juntingnet_diag}. The network consists of three stages of CONV-ReLU-MAX\_POOL followed by two fully-connected layers, the last of which is subjected to a maxout operation. The input to the network is a $128 \times 128$ RGB image and the output is a $1024$ dimensional vector that is resized to a $32 \times 32$ salience map. This is a 1024-D regression task with element wise squared error loss.  Receptive fields of [5$\times$5], [3$\times$3] and [3$\times$3] are used in 1$^{st}$, 2$^{nd}$ and 3$^{rd}$ stage on convolution. Receptive field of MAX\_POOL layer is [2$\times$2].
 During pre-training, networks weights were initialized by uniform sampling from a zero mean Normal distribution with standard deviation of 0.01. The bias terms were set to 0.1 at beginning. We used stochastic gradient descent with Nestorov momentum for faster convergence.The learning rate was adaptively decreased from 0.3 at beginning to 10$^{-4}$ at end of training. Upon culmination of training, 32$\times$32 map later resized to exact resolution of input image using bilinear interpolation.

\section {Two step Hierarchical Learning}
In this section we describe our proposed two step hierarchical framework for training a CNN for visual saliency prediction task.
\subsection{Weakly Pretraining Stage on ImageNet}
In supervised learning paradigm, weak training is a neoteric attempt of reducing human effort for meticulously creating enormous ground truth dataset for large scale learning frameworks. The key idea is to extract auxiliary information from unannotated data. ImageNet \cite{imagenet}, for example, has about 1 million natural images. For generating ground truth eye fixations on ImageNet, a human operator has to look over the entire dataset; such task is definitely not prudent.

\begin{figure*}
	\centering
	\subfigure {\includegraphics[scale=.25]{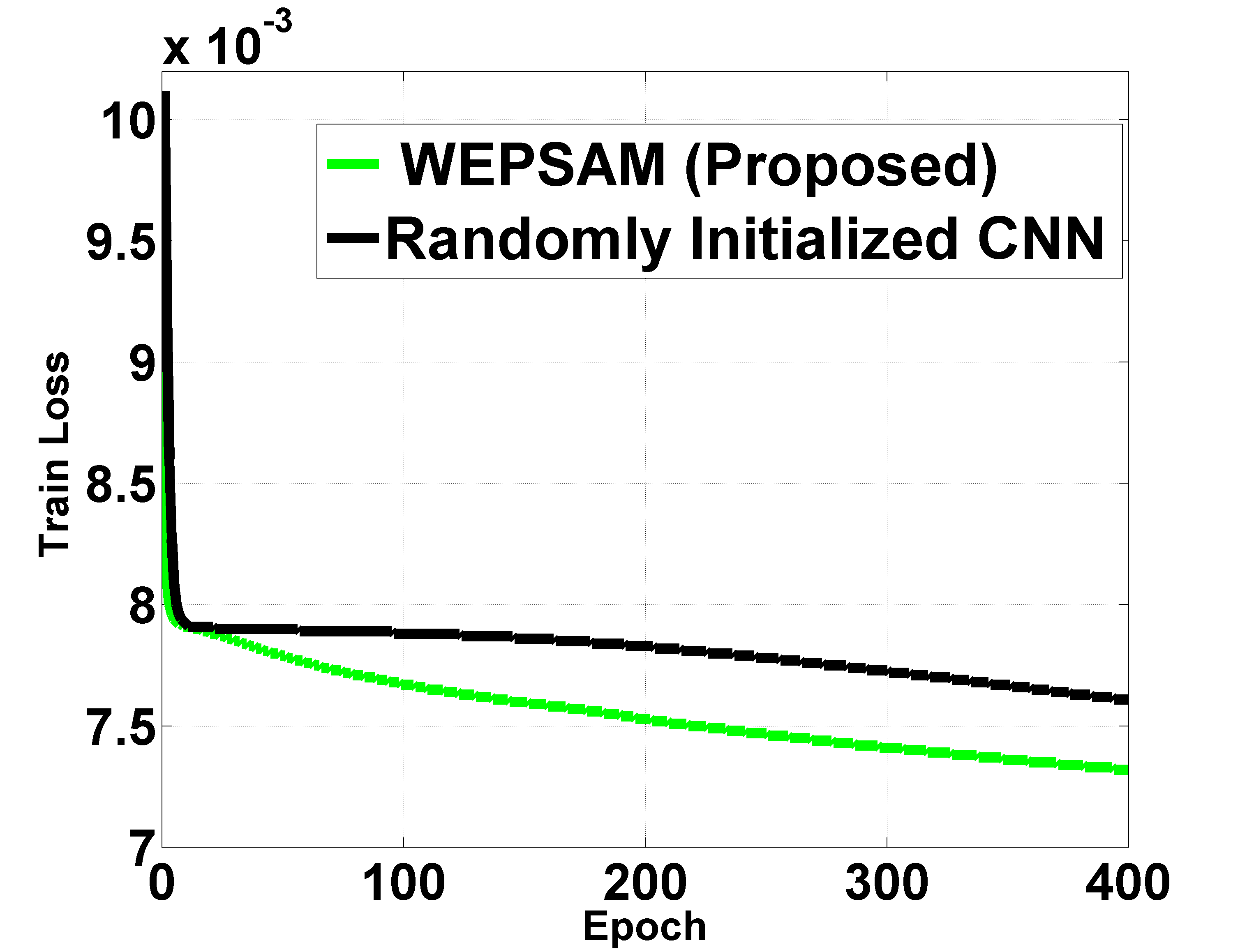}}
	\subfigure {\includegraphics[scale=.25]{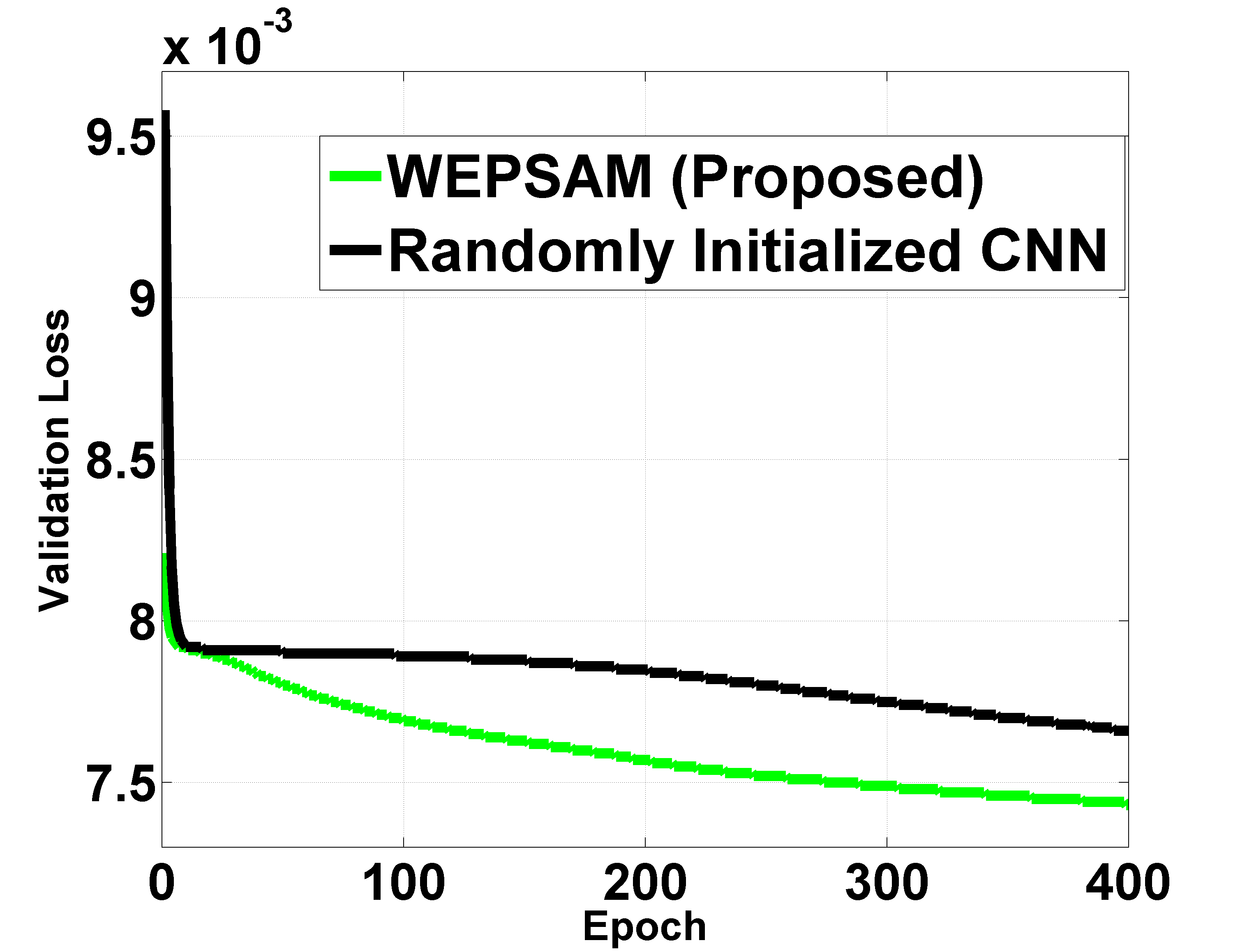}}\\
	\caption{Training and validation loss of training a CNN model on human eye fixation maps of iSUN and SALICON datasets. Loss is defined as the average of pixelwise squared error between ground truth saliency map and predicted map. It is evident that a weakly pretrained model such as WEPSAM fosters faster convergence rate.}
	\label{fig_pretrain_compare}
\end{figure*}

 We propose an elegant solution to circumnavigate this problem. Our work is motivated by the fact that supervised pre-training followed by fine tuning fosters faster convergence rate in CNN\cite{pre_train}. Each RGB image is first down sampled to 128X128X3 and then we create a gray scale saliency map of 32X32 using a graph based saliency model \cite{gbvs}. It is to be noted that the maps produced by \cite{gbvs} only provide an approximation to actual saliency prediction behavior of human visual system. Saliency maps produced by \cite{gbvs} tend to much more diffused compared to actual eye fixation, specially in scarcity of salient objects in an image. We define a filter criterion based on entropy of the maps. Entropy of an image $I(x,y)$ is defined as,$E=-\sum_{i=1}^{256} p_i \log _2 {p_i}$
where, $p_i$ denotes the normalized histogram count of $i^{th}$ bin. To imitate human eye fixation, it is desired to generate low entropy saliency maps. So, for pre-training, we sort the maps according to increasing order of entropy and select the top 10$^5$ entries for pre-training the CNN. We pre-train the CNN model for 500 epochs.

\subsection{Fine Tuning of Weak Model}
\vspace*{-3mm}In this stage we use actual ground truth fixations from widely used public databases for fine tuning our previously developed weak trained CNN model. In this stage, we use the same CNN architecture but initialize the network with weights learnt in pre-training step. This ensures that we achieve faster rate of error convergence on training set and simultaneously manifest better generalization performance. Training in this stage has been run for 1200 epochs, after which, both training and validation loss begin to saturate.  

\begin{table}
	\scriptsize
	\begin{center}
		\begin{tabular}{|l|c|c|c|c|c|c|c|}
			\hline
			Model& AUC-Judd&AUC-Borji&CC&SIM&KL&NSS\\
			\hline\hline
			MR-CNN\cite{mrcnn}& 0.79&0.75&0.48&\textbf{0.48}&1.08&\textbf{1.37} \\
			CNN-VLM\cite{kato}& 0.79&\textbf{0.79}&0.44&0.43&1.06&1.18 \\
			MKL\cite{kavak}& 0.78&0.78&0.42&0.42&1.10&1.08 \\
			RARE-2012\cite{Riche}& 0.77&0.75&0.42&0.46&\textbf{1.01}&1.15 \\
			CAS\cite{context}& 0.74&0.73&0.36&0.43&1.06&0.95\\
			LGS\cite{lgs}& 0.76&0.76&0.39&0.42&1.11 &1.02\\
			GNMS\cite{Zhong} & 0.74&0.67&0.34&0.42&1.21&0.97 \\
			NARFI\cite{narfi}& 0.73&0.61&0.31&0.38&5.17&0.83 \\
			STC\cite{Holzbach} & 0.79&0.78&0.40&0.39&1.23&0.97 \\
			CIW\cite{Murray} & 0.70&0.69&0.27&0.38&1.23&0.73 \\
			\textbf{WEPSAM}&\textbf{0.80}&0.78&\textbf{0.51}&0.45&\textbf{1.01}&1.35 \\
			\textbf{(Proposed)}&&&&&&\\
			\hline
		\end{tabular}
		\label{table_r}
	\end{center}
	\caption{Quantitative comparison between different saliency models on the challenging MIT300 dataset. Best results are marked in bold. Though MR-CNN has slightly better SIM and NSS metric compared to WEPSAM, complexity of MR-CNN is much higher because it trains 3 CNNs at multiple scales. Also, the basic CNN architecture of MR-CNN is more complex compared to WEPSAM.}
	\label{table_metric_comparison} 
\end{table}
\begin{figure*}
	\centering
	\subfigure {\includegraphics[scale=.08]{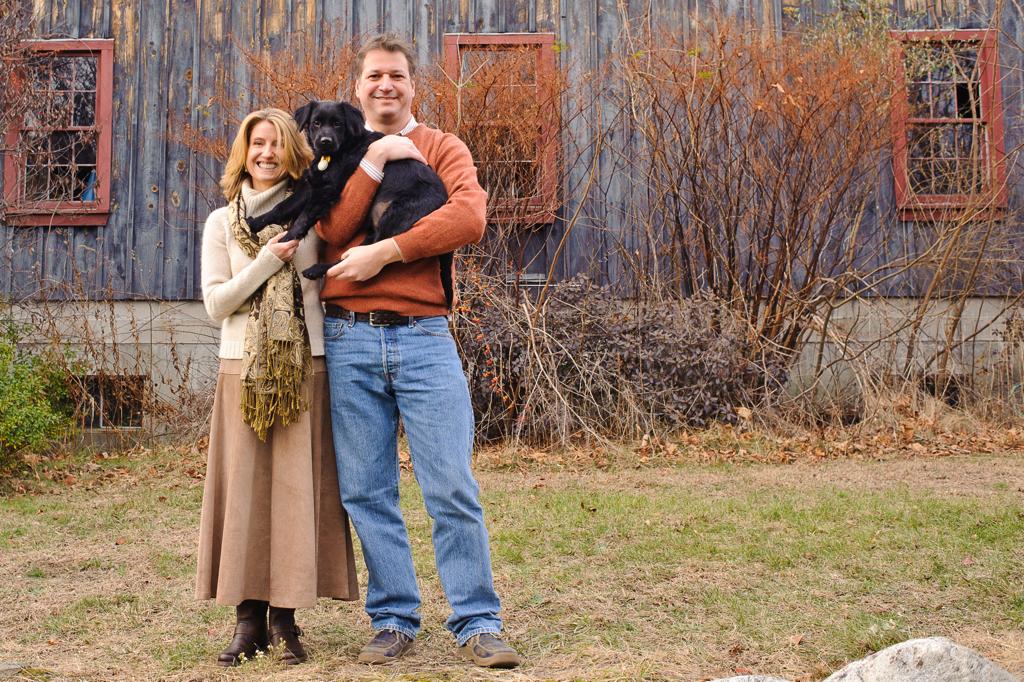}}
	\subfigure {\includegraphics[scale=.08]{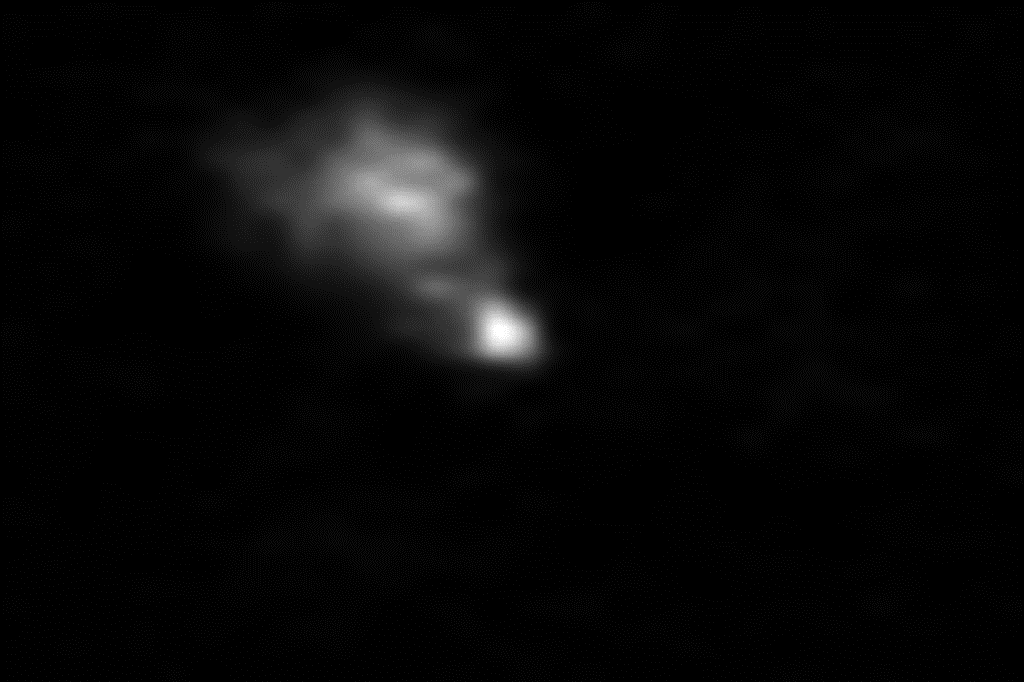}}
	\subfigure {\includegraphics[scale=.08]{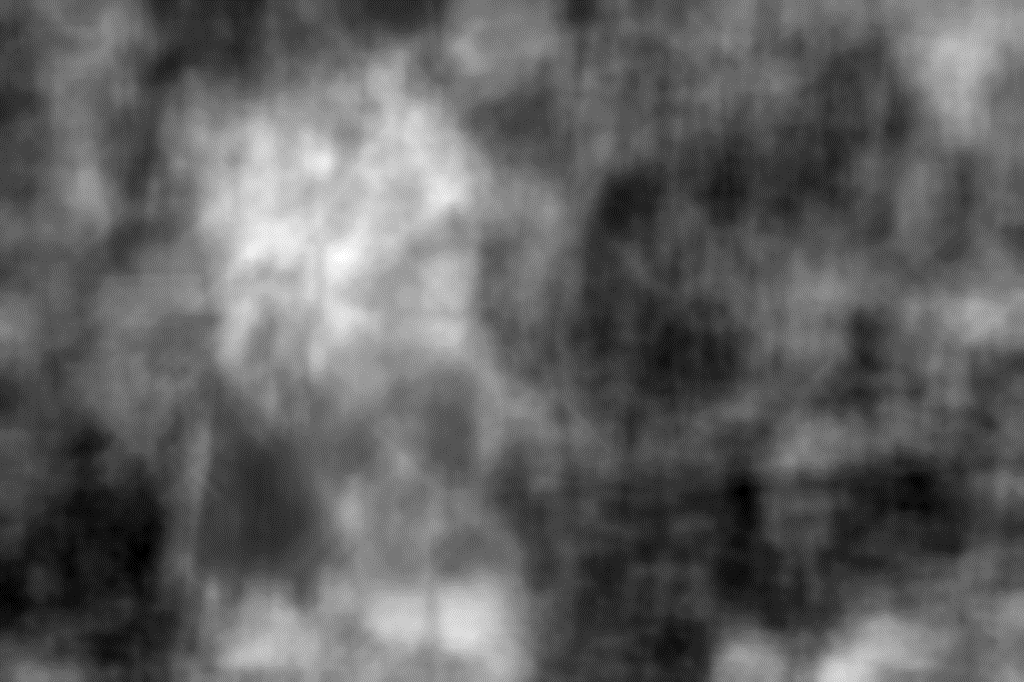}}
	\subfigure {\includegraphics[scale=.08]{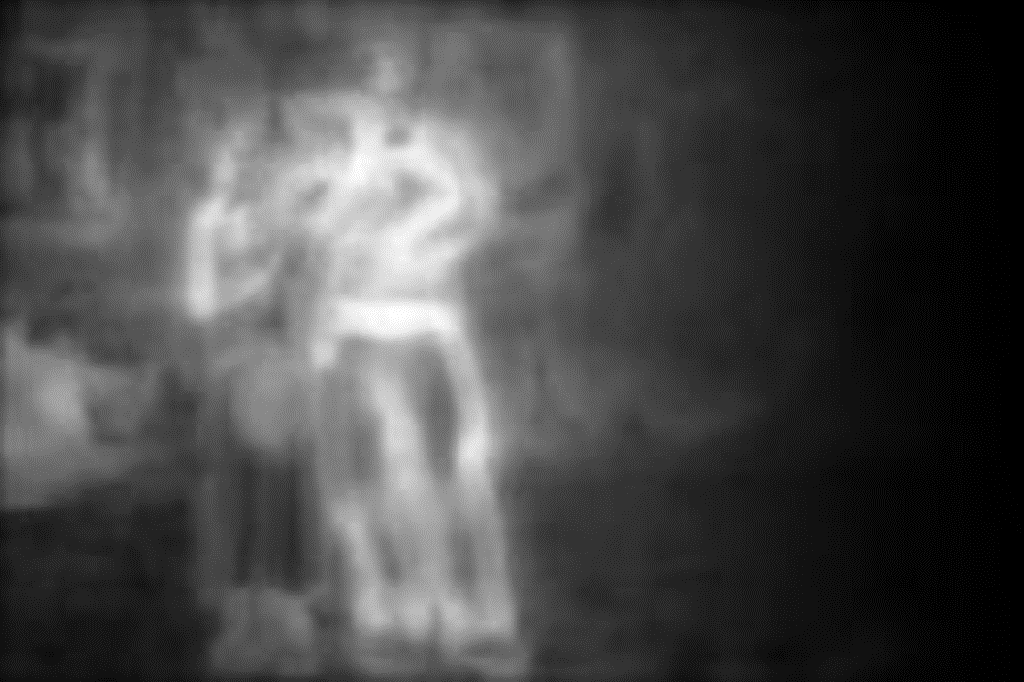}}
	\subfigure {\includegraphics[scale=.08]{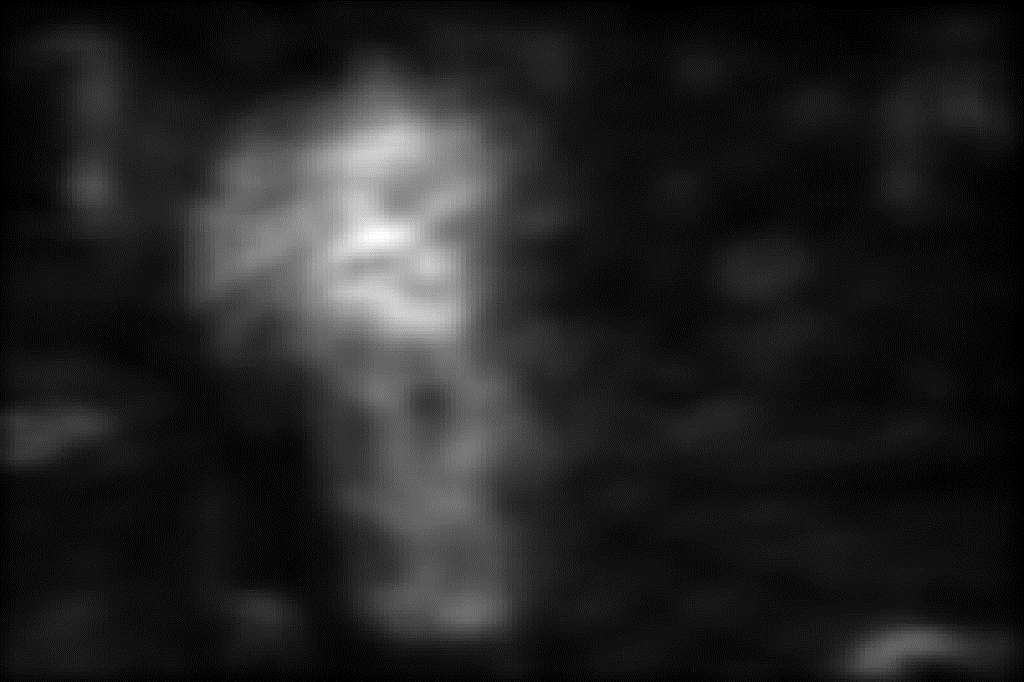}}\\
\vspace{-0.6em}	
	\subfigure {\includegraphics[scale=.08]{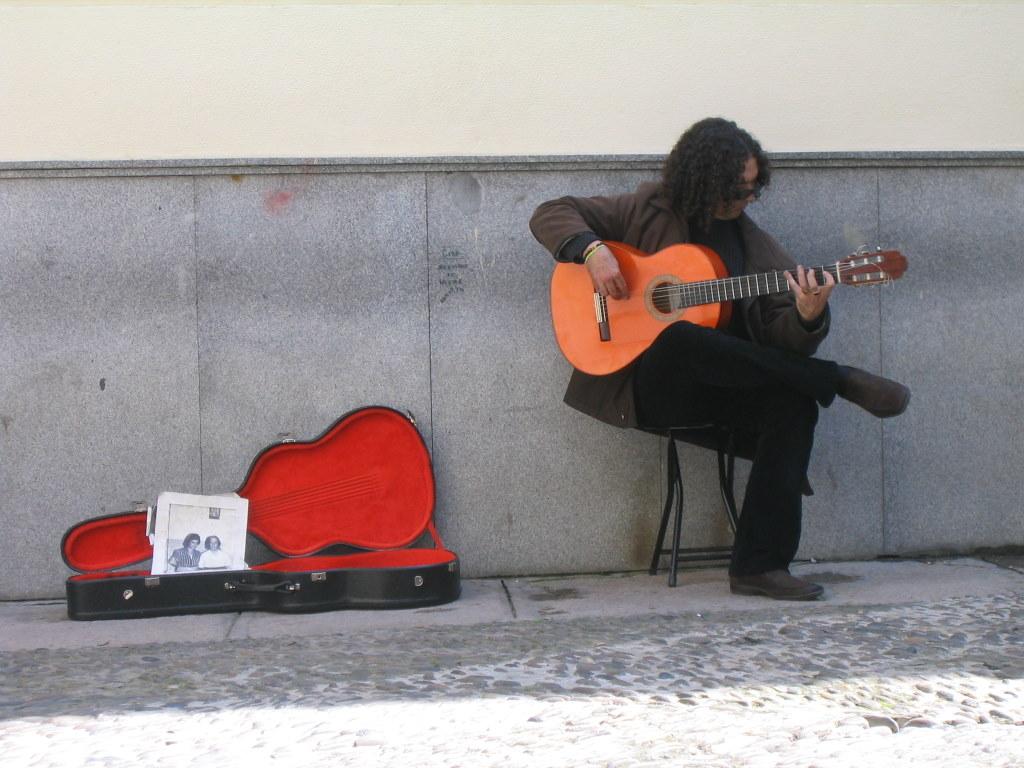}}
	\subfigure {\includegraphics[scale=.08]{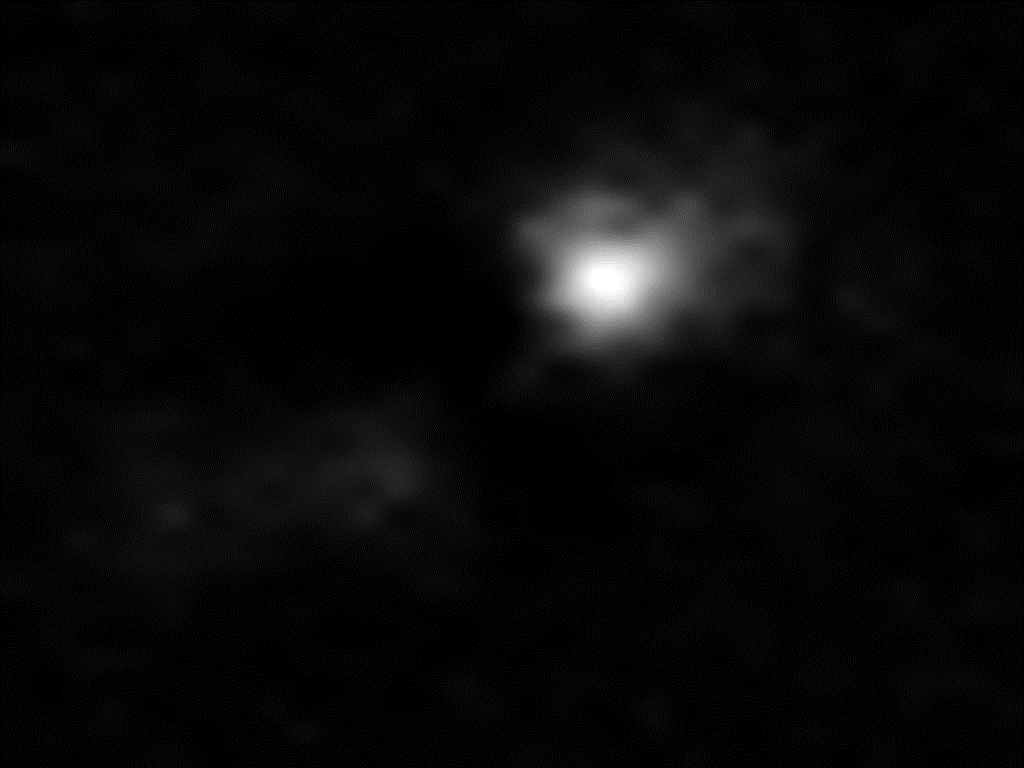}}
	\subfigure {\includegraphics[scale=.08]{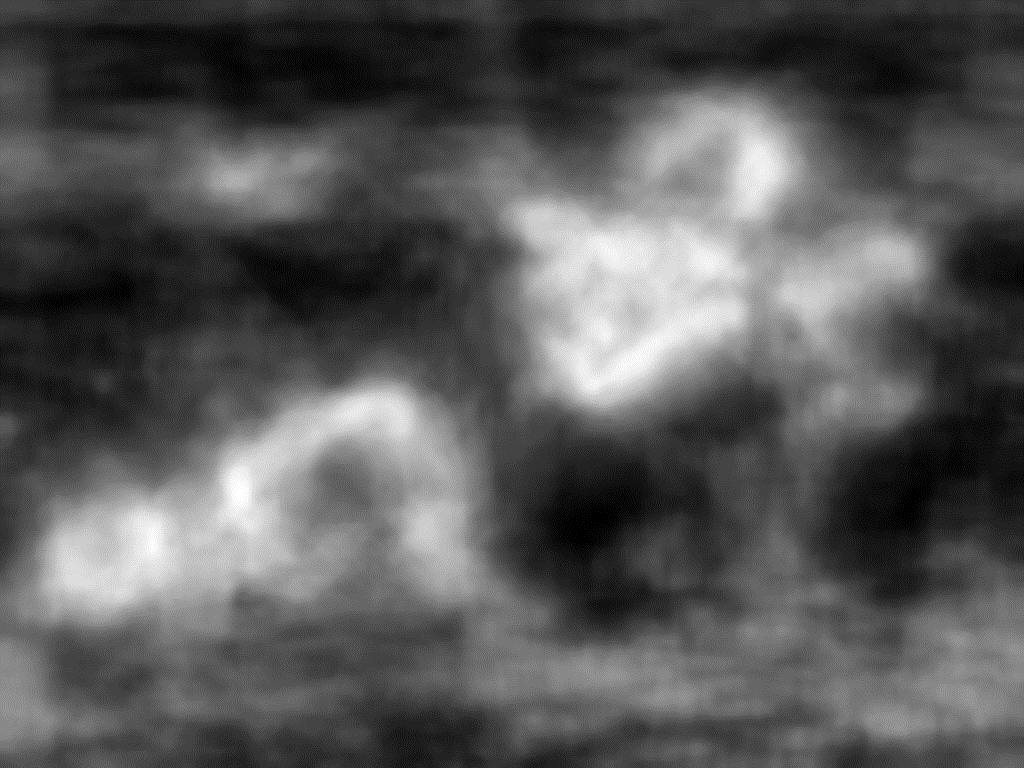}}
	\subfigure {\includegraphics[scale=.08]{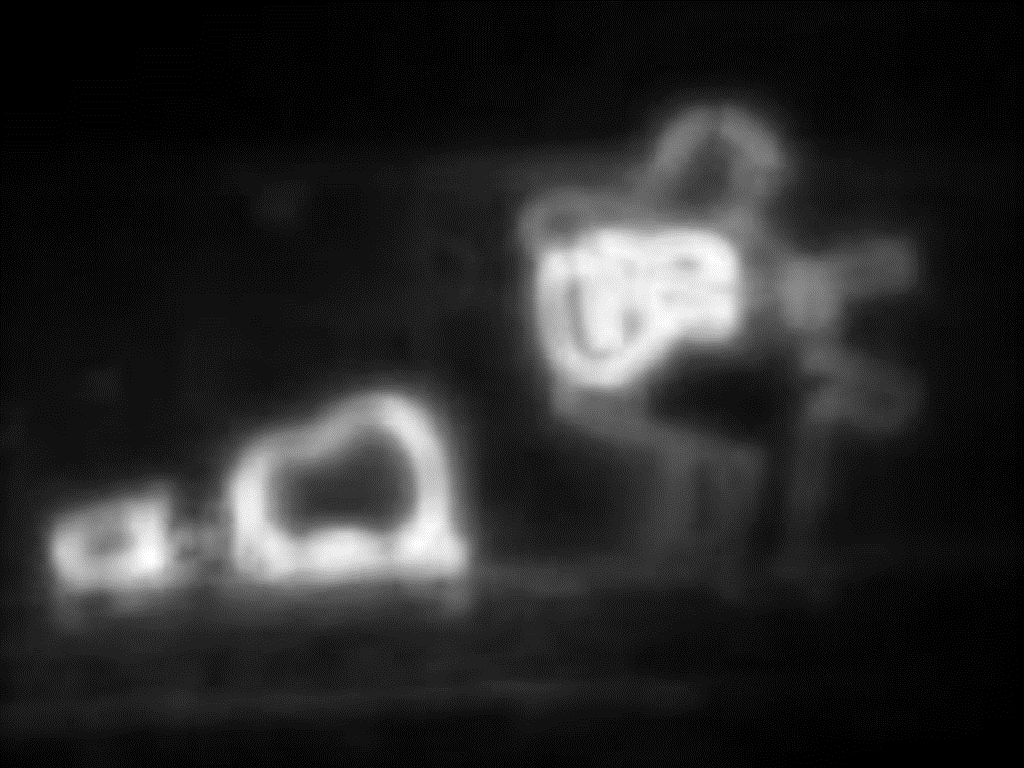}}
	\subfigure {\includegraphics[scale=.08]{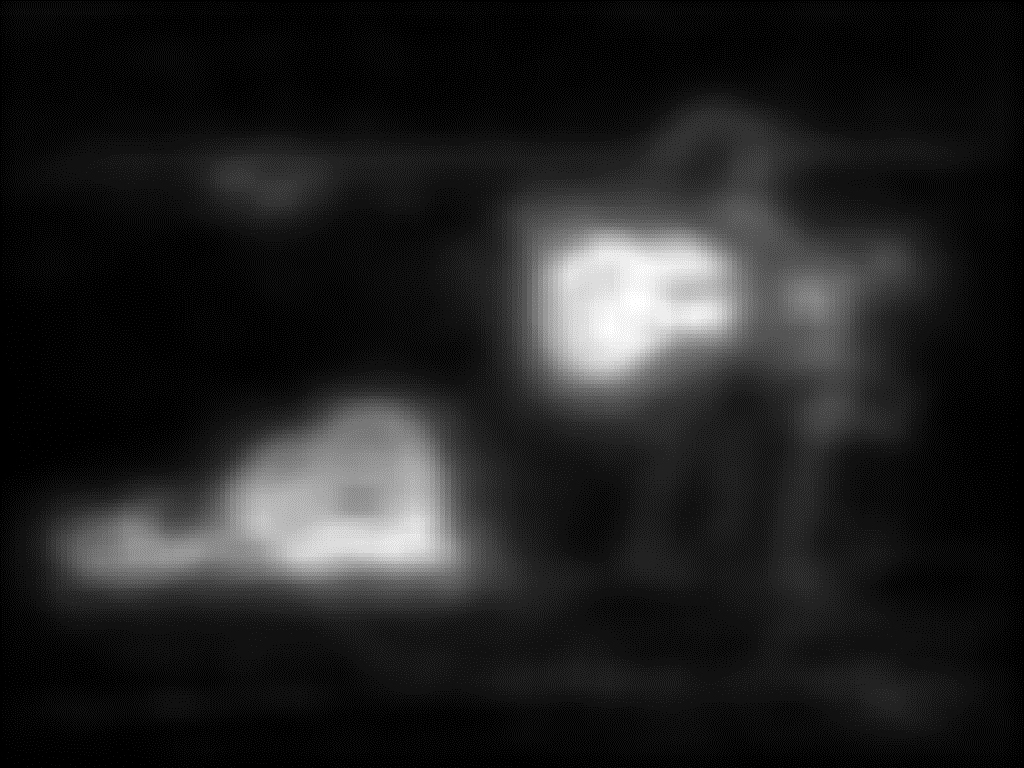}}\\
\vspace{-0.6em}	
	\subfigure {\includegraphics[scale=.08]{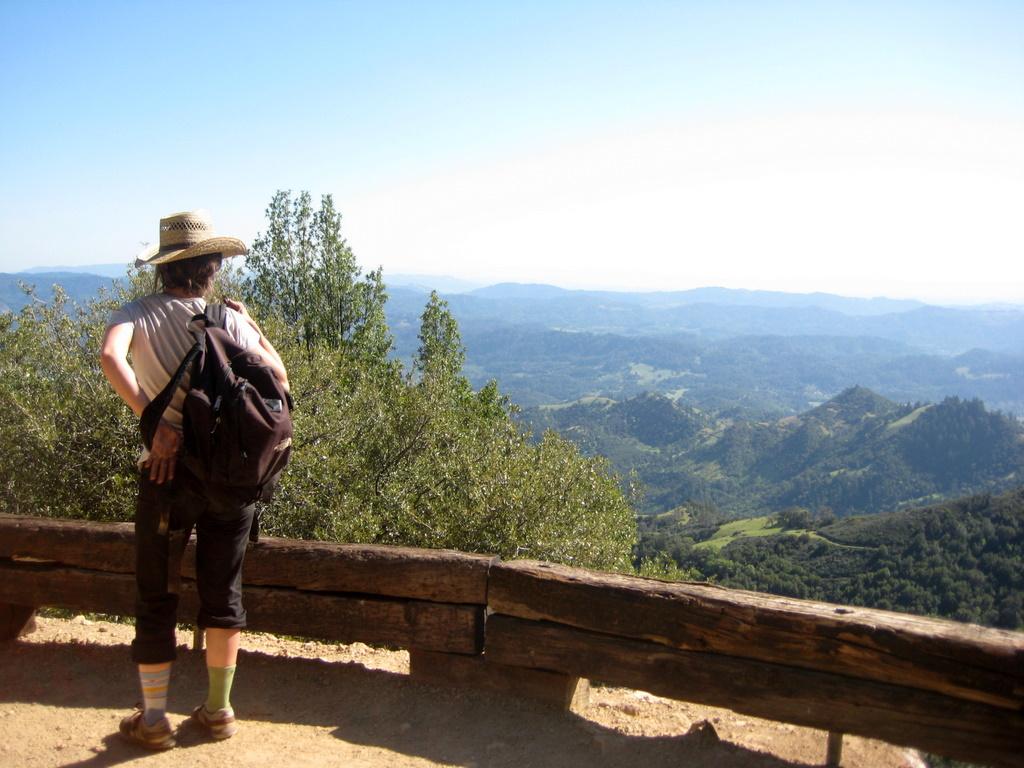}}
	\subfigure {\includegraphics[scale=.08]{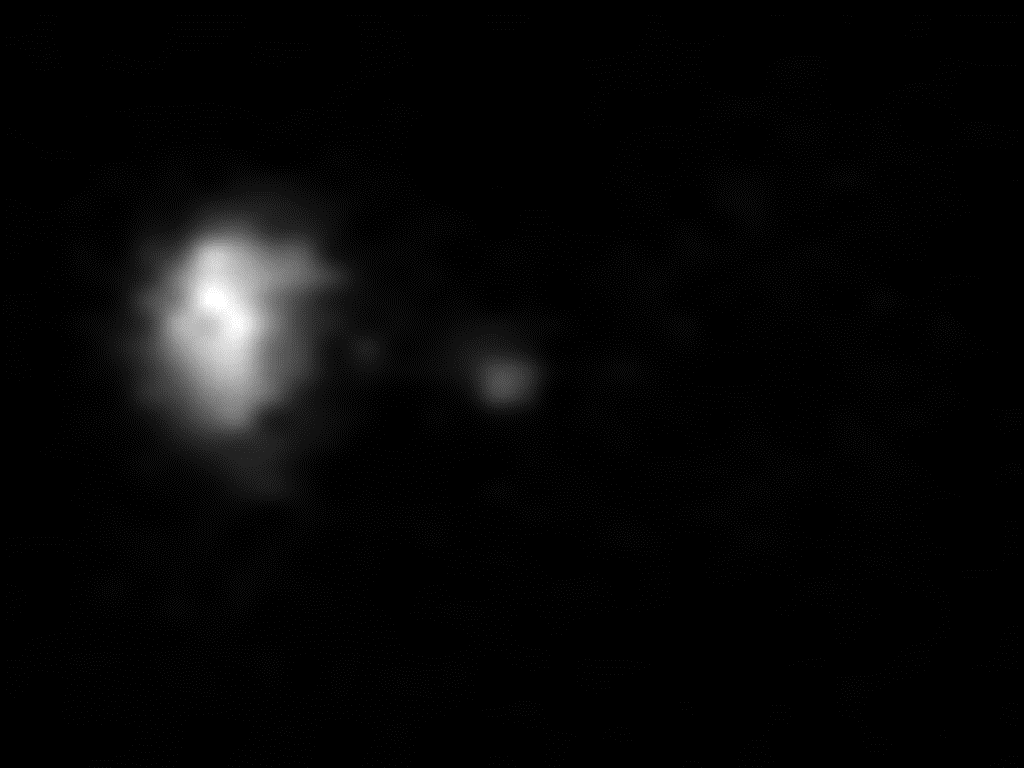}}
	\subfigure {\includegraphics[scale=.08]{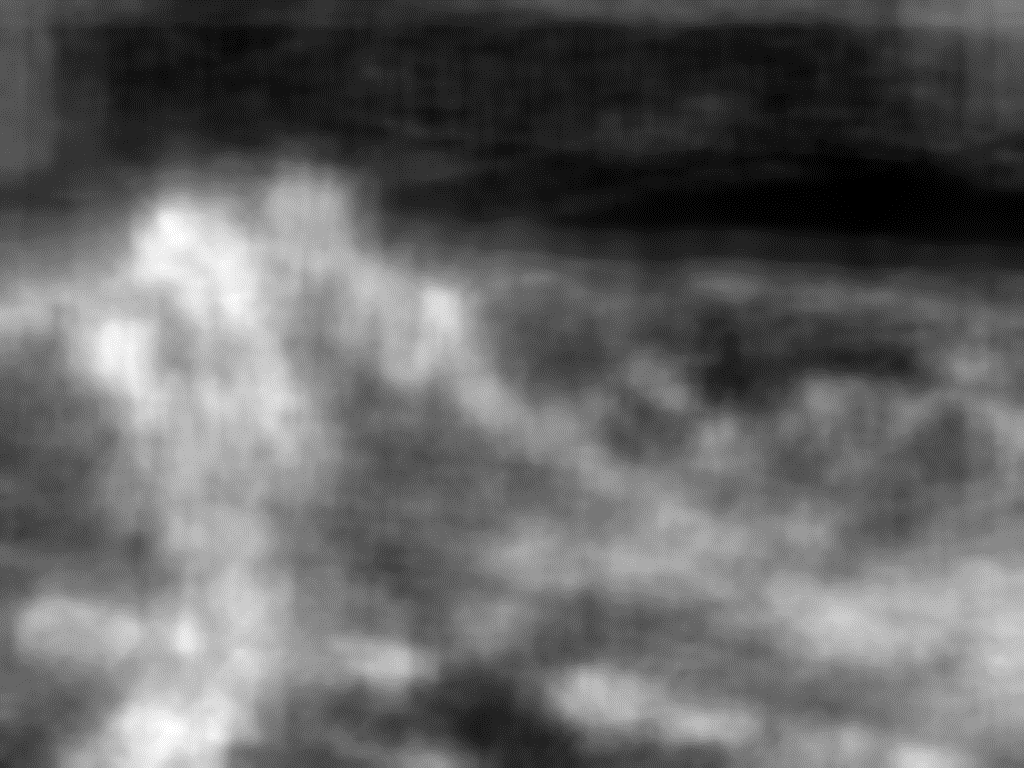}}
	\subfigure {\includegraphics[scale=.08]{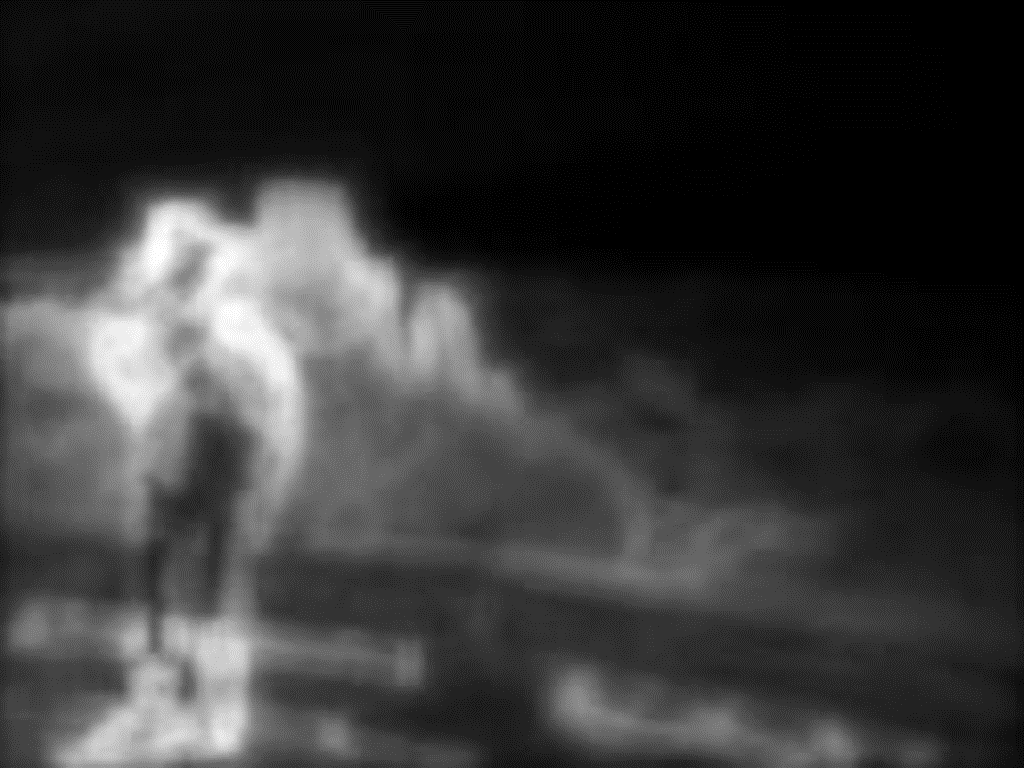}}
	\subfigure {\includegraphics[scale=.08]{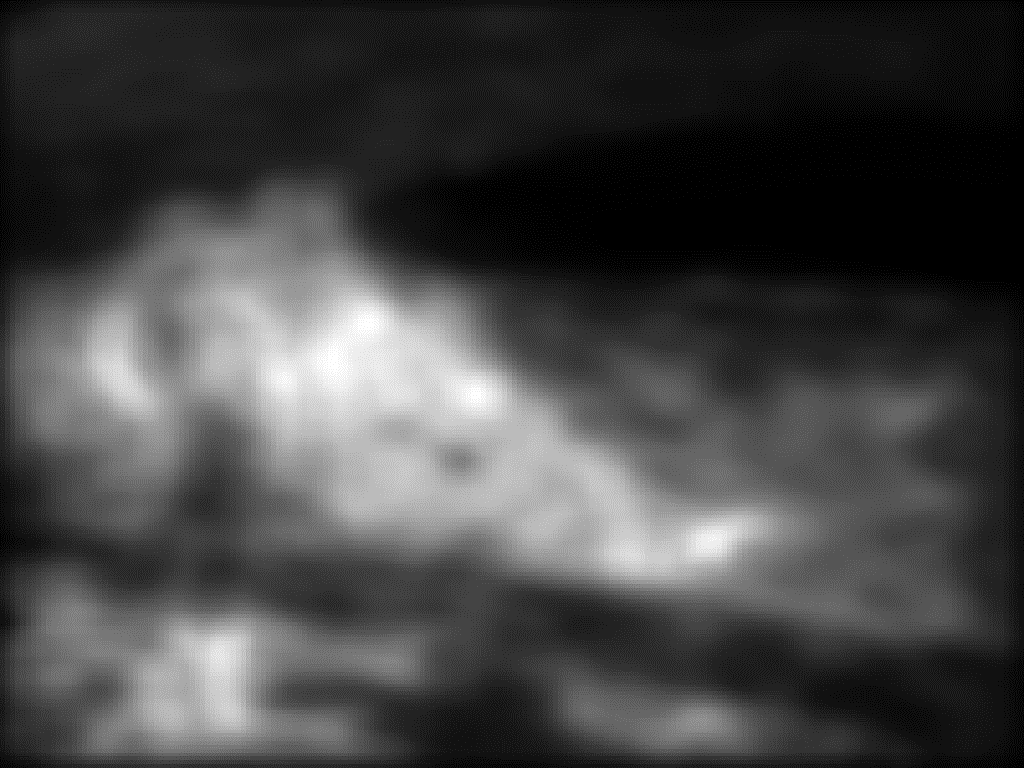}}\\	
	
	\caption{Comparsion of saliency maps on some exemplary MIT300 test set. \textbf{Col 1:} Actual RGB image, \textbf{Col 2:} Saliency map by our proposed WEPSAM, \textbf{Col 3:} CIW \cite{Murray}, \textbf{Col 4:} CAS \cite{context}, \textbf{Col 5:} RARE-2012 \cite{Riche} . Our proposed model emphasizes only those regions in an image where a human would look at first glance. Competing models, instead, highlight mainly the edges and thereby produces much more diffused map. WEPSAM model is thus superior in identifying semantically important image regions.}
	\label{fig_map_compare}
\end{figure*}
\section{Results and Discussions}
\vspace{-5mm} In this section we demonstrate experimental results to evaluate the efficacy of the proposed approach. In the first part of our results, we show how pre-training using weak data helps us to train the convolutional network faster. In the second part, we test our model on the challenging MIT300 dataset and compare its performance quantitatively and qualitatively with recent state of the art methods. For fine tuning our network after pre-training we have used ground truths from iSUN \cite{iSUN_dataset} and SALICON \cite{salicon_dataset} datasets. iSUN contains 6000 training and 926 validation image-map pairs. SALICON dataset has 10000 training and 5000 validation pairs.
 \subsection{Effect of Pre-Training}
 In Fig. \ref{fig_pretrain_compare} we plot the training and validation loss on the combined ground truth maps of iSUN and SALICON datsets. The green lines show the train and validation loss for proposed WEPSAM model while black denotes the metrics for a randomly initialized net of same architecture.

  It is evident that pre-training fosters faster decay of training loss compared to a randomly initialized net and simultaneously manifests better generalization accuracy.Specifically at onset of training, train and validation loss for WEPSAM are 8.2$\times$10$^{-3}$ and 8.4$\times$10$^{-3}$ respectively, while those of random initialized net are 10.6$\times$10$^{-3}$ and 9.6$\times$10$^{-3}$. After 400 epochs, train and validation loss for WEPSAM are 7.2$\times$10$^{-3}$ and 7.3$\times$10$^{-3}$ respectively, while those of random initialized net are 7.7$\times$10$^{-3}$ and 7.8$\times$10$^{-3}$. During weak pre-training weights of our network were learnt so as to approximately imitate human eye fixation model. Thus, during fine tuning, prediction of pre-learnt net is much more coherent with ground truth than a randomly initialized net. 
\subsection{Performance on MIT300 Database}
Next we compare saliency prediction performance on the challenging MIT300 dataset \cite{mit_dataset1,mit_dataset2}. It is to be noted that WEPSAM was fine-tuned only on images of iSUN and SALICON datasets and thus the test images are substantially different than training images. We compare our model with recent state-of-the-art methods such as multi resolution CNN (MR-CNN) \cite{mrcnn}, CNN-VLM \cite{kato}, multiple kernel based learning (MKL)\cite{kavak}, RARE-2012\cite{Riche}, Context Aware Saliency Model(CAS) \cite{context}, Local+Global Saliency Model (LGS) \cite{lgs}, Generalized Nonlocal Mean Saliency (GNMS) \cite{Zhong}, NARFI saliency (NARFI) \cite{narfi}, Sampled Template Collation (STC) \cite{Holzbach} and Chromatic Induction Wavelet Model (CIW) \cite{Murray}. The first three models are essentially learning based. In Table \ref{table_metric_comparison} we compare the performances of competing models based on six popularly used metrics, viz., AUC-Judd, AUC-Borji, CC (correlation coefficient), SIM (similarity metric), KL (Kullback- Leibler divergence) and NSS (normalized scanpath saliency). From Table \ref{table_metric_comparison} we see that proposed WEPSAM outperforms non learning based methods by significant margins on multiple metrics. MR-CNN outperforms our model on SIM and KL but it is to be noted that MR-CNN model is much more complex than WEPSAM. MR-CNN trains three different CNNs on [400 $\times$ 400], [250 $\times$ 250] and [150 $\times$ 150] scales with 6 layers of convolution. In contrast proposed WEPSAM only uses a single resolution of [128 $\times$ 128] using only 3 layers of convolution. In Fig. \ref{fig_map_compare} we present the saliency maps on three images of MIT300. Ground truths have not been released to public but intuitively we can see that WEPSAM emphasizes only those regions in an image which are semantically important to a human. Competing methods mainly highlight the image gradients and thereby manifesting diffused, semantically insignificant maps.

\section{Conclusion}
In this paper we presented WEPSAM as a pioneering effort of developing a weakly pre-trained end-to-end CNN based model for saliency prediction. WEPSAM used an elegant approach of weakly learning saliency maps on ImageNet. Such pre-training acted as a regularizer and fostered in quicker convergence of validation loss on ground truth eye fixations. We hope that this work will instigate a new genre of research of using auxiliary data for saliency modeling. In future, we wish to test our model with more complex CNN models such as GoogleNet \cite{googlenet} and VGGnet \cite{vggnet} to exploit the benefit of pre-learning on a larger scale.
{\small
\bibliographystyle{IEEEbib}
\bibliography{refs}}

\end{document}